\documentclass[11pt,a4paper]{article}
\pdfoutput=1
\usepackage[hyperref]{emnlp2018}
\usepackage{times}
\usepackage{latexsym}
\usepackage{graphicx}
\usepackage{url}
\usepackage{multirow}
\usepackage[inline]{enumitem}
\usepackage{booktabs}
\usepackage{xcolor}
\usepackage{colortbl}
\usepackage{xspace}
\usepackage{bbm}
\usepackage{amsfonts}
\usepackage{amsmath}
\usepackage[linesnumbered,ruled,vlined]{algorithm2e}

\SetCommentSty{mycommfont}

\usepackage{ifthen}
\newcolumntype{g}{>{\columncolor{black!5}}c}
\newcolumntype{f}{>{\columncolor{black!5}}r}
\newcolumntype{L}[1]{>{\columncolor{black!5}}m{#1}}

\aclfinalcopy 
\def\aclpaperid{2164} 

\usepackage{amsmath}
\DeclareMathOperator*{\argmax}{arg\,max}
\usepackage{verbatim,toappendix}

\newcommand{\hal}[1]{}
\newcommand{\kathy}[1]{}
\newcommand{\hidey}[1]{}

\title{Content Selection in Deep Learning Models of Summarization}

\author{Chris Kedzie \and Kathleen McKeown\\
  Department of Computer Science \\
  Columbia University \\
  {\tt \{kedzie,kathy\}@cs.columbia.edu} \\\And
  Hal Daum\'e III  \\
  University of Maryland, College Park \\
  Microsoft Research, New York City \\
  {\tt hal@cs.umd.edu} \\}

\date{}

\begin{document}
\maketitle
\begin{abstract}
We carry out experiments with deep learning models of summarization across the domains of news, personal stories, meetings, and medical articles in order to understand how content selection is performed. We find that many sophisticated features of state of the art extractive summarizers do not improve performance over simpler models. These results suggest that it is easier to create a summarizer for  a new domain than previous work suggests and bring into question the benefit of deep learning models for summarization for those domains that do have massive datasets (i.e., news). At the same time, they suggest important questions for new research in summarization; namely,  new forms of sentence representations or external knowledge sources are needed that are better suited to the summarization task. 
\end{abstract}

\newcommand{\rouge}{\operatorname{ROUGE}}

\newcommand{\sent}[1][]{%
  \ifthenelse{ \equal{#1}{} }
     {\ensuremath{s}}
     {\ensuremath{s_{#1}}}
}
\newcommand{\sentEmb}[1][]{%
  \ifthenelse{ \equal{#1}{} }
     {\ensuremath{h}}
     {\ensuremath{h_{#1}}}
}
\newcommand{\sentEmbSize}{\ensuremath{d}}

\newcommand{\docSize}{\ensuremath{n}}

\newcommand{\slabel}[1][]{%
  \ifthenelse{ \equal{#1}{} }
     {\ensuremath{y}}
     {\ensuremath{y_{#1}}}
}

\newcommand{\wordEmb}[1][]{%
  \ifthenelse{ \equal{#1}{} }
     {\ensuremath{w}}
     {\ensuremath{w_{#1}}}
}

\newcommand{\sentSize}{\ensuremath{|\sent|}}
\newcommand{\wordEmbSize}{\ensuremath{{n^\prime}}}
\newcommand{\rSentEmb}[1][]{
  \ifthenelse{ \equal{#1}{} }
  {\ensuremath{\overrightarrow{\sentEmb}}}
  {\ensuremath{\overrightarrow{\sentEmb}_{#1}}}
}
\newcommand{\lSentEmb}[1][]{
  \ifthenelse{ \equal{#1}{} }
  {\ensuremath{\overleftarrow{\sentEmb}}}
  {\ensuremath{\overleftarrow{\sentEmb}_{#1}}}
}

\newcommand{\doc}{d}
\newcommand{\docsize}{d}
\newcommand{\sentvec}{h}
\newcommand{\summary}{S}
\newcommand{\wemb}{w}
\newcommand{\wembdim}{n}

\newcommand{\rSentVec}{\overrightarrow{\sentvec}}
\newcommand{\lSentVec}{\overleftarrow{\sentvec}}

\newcommand{\numFeatureMaps}{m}
\newcommand{\maxFeatureMaps}{M}
\newcommand{\maxWindowSize}{K}
\newcommand{\filterWindowSize}{k}
\newcommand{\mapsSize}{(\numFeatureMaps,\filterWindowSize)}
\newcommand{\genActivation}{a}
\newcommand{\specActivation}{\genActivation^{\mapsSize}}
\newcommand{\genConvWeight}{W}
\newcommand{\genConvBias}{b}
\newcommand{\specConvBias}{\genConvBias^{\mapsSize}}
\newcommand{\specConvWeight}{\genConvWeight^{\mapsSize}}
\newcommand{\genFeatureMap}{h}
\newcommand{\specFeatureMap}{\genFeatureMap^{\mapsSize}}
\newcommand{\relu}{\operatorname{ReLU}}
\newcommand{\gru}{\operatorname{GRU}}
\newcommand{\rgru}{\overrightarrow{\gru}}
\newcommand{\lgru}{\overleftarrow{\gru}}
\newcommand{\extHidden}{z}
\newcommand{\rExtHidden}{\overrightarrow{\extHidden}}
\newcommand{\lExtHidden}{\overleftarrow{\extHidden}}
\newcommand{\senc}{\operatorname{enc}}
\newcommand{\explicitLikelihood}{p(\slabel_1,\ldots,\slabel_{\sentSize}|
                                   \sentvec_1, \ldots, \sentvec_{\sentSize})}
\newcommand{\compactLikelihood}{p(\slabel|\sentvec)}
\newcommand{\naiveLikelihood}{\prod_{i=1}^{\sentSize}p(\slabel_i|
                              \sentvec_1, \ldots, \sentvec_{\sentSize})}
\newcommand{\markovLikelihood}{\prod_{i=1}^{\sentSize} 
         p(\slabel_i|\slabel_{<i}, \sentvec_1, \ldots, \sentvec_{\sentSize})}

\newcommand{\logits}{a}

\newcommand{\sts}{Seq2Seq}
\newcommand{\stsbf}{\textbf{\sts}}

\newcommand{\baselineOne}{C\&L\xspace}
\newcommand{\baselineOneBF}{\textbf{\baselineOne}\xspace}
\newcommand{\baselineTwo}{SR\xspace}
\newcommand{\baselineTwoBF}{\textbf{\baselineTwo}\xspace}

\newcommand{\modelOne}{RNN\xspace}
\newcommand{\modelOneBF}{\textbf{\modelOne}\xspace}
\newcommand{\modelTwo}{Seq2Seq\xspace}
\newcommand{\modelTwoBF}{\textbf{\modelTwo}\xspace}

\newcommand{\encExtHidden}{z}
\newcommand{\rEncExtHidden}{\overrightarrow{\encExtHidden}}
\newcommand{\lEncExtHidden}{\overleftarrow{\encExtHidden}}
\newcommand{\decExtHidden}{q}
\newcommand{\rDecExtHidden}{\overrightarrow{\decExtHidden}}
\newcommand{\lDecExtHidden}{\overleftarrow{\decExtHidden}}
\newcommand{\attnExtHidden}{\bar{\encExtHidden}}

\section{Introduction}

Content selection is a central component in many natural language generation
tasks,
where, given a generation goal, the system must determine which information
should be expressed in the output text \cite{gatt2018survey}.
In summarization, content selection is usually accomplished through sentence (and,
occasionally, phrase) extraction.
Despite being a key component of both
extractive and abstractive summarization systems, it is is not well
understood how deep learning models perform content selection with only word and 
sentence
embedding based features as input.
Non-neural network approaches often use frequency and information theoretic measures as proxies
for content salience \cite{hong2014improving}, but these are not explicitly 
used in most neural network summarization systems.


In this paper, we seek to better understand how deep learning models of 
summarization perform content selection across multiple domains (\S~\ref{sec:datasets}): news, personal stories,
meetings, and medical articles (for which we collect a new corpus).\footnote{Data preprocessing and implementation code can be found here: \url{https://github.com/kedz/nnsum/tree/emnlp18-release}}
We analyze
several recent sentence extractive neural network architectures, 
specifically considering the design choices for sentence encoders (\S~\ref{sec:senc})
and sentence extractors (\S~\ref{sec:sext}). We compare Recurrent Neural Network (RNN) and Convolutional Neural
Network (CNN) based sentence representations to the 
simpler approach of word embedding averaging to understand the gains 
derived from more sophisticated architectures.
We also question the necessity of auto-regressive sentence extraction 
(i.e. using previous predictions to inform future predictions), 
which previous approaches have used (\S~\ref{sec:related}),
and propose two alternative models that extract sentences independently.
%
%
%
%
\\[-0.5em]

\noindent
Our main results (\S~\ref{sec:exps}) reveal:
\begin{enumerate}[noitemsep]
\item Sentence position bias dominates the learning signal for news summarization, though not for
    other domains.\footnote{This is a known bias 
    in news summarization \cite{nenkova2005automatic}.}
Summary quality for news is only slightly degraded when content words
are omitted from sentence embeddings. 
\item Word embedding averaging is as good or better than either RNNs or CNNs for sentence embedding across all domains.
\item Pre-trained word embeddings are as good, or better than, learned embeddings in five of six datasets.
\item Non auto-regressive sentence extraction performs as good or better 
     than auto-regressive extraction in all
    domains.
\end{enumerate} 

\noindent
Taken together, these and other results in the paper suggest that we are 
over-estimating the ability of deep learning models to learn robust and 
meaningful content features for summarization.  
In one sense, this might lessen the burden of applying neural network models
of  content to other domains; one really just needs in-domain word embeddings.
However, if we want to learn something other than where the start of 
the article is, we will need to design other means of sentence representation,
and possibly external knowledge representations, better suited to the summarization task.


\section{Related Work} \label{sec:related}

The introduction of the CNN-DailyMail corpus by \citet{nips15_hermann} 
allowed for the application of large-scale training of deep learning models 
for summarization.
\citet{cheng2016neural} developed a sentence extractive model that uses a 
word level CNN to encode sentences and a sentence level sequence-to-sequence 
model to predict which sentences to include in the summary. Subsequently, 
\citet{nallapati2017summarunner} proposed a different model using word-level 
bidirectional RNNs along with a sentence level 
bidirectional RNN for predicting which sentences should be extracted. 
Their sentence extractor creates representations of the whole document and 
computes separate scores for salience, novelty, and location.
These works represent the state-of-the-art for deep learning-based extractive
summarization and we analyze them further in this paper.

Other recent neural network approaches include, \citet{yasunaga2017graph},
who learn a graph-convolutional network (GCN) for multi-document summarization.
They do not 
closely examine the choice of sentence encoder, which is one of the focuses
of the present paper; rather, they study the best choice of graph 
structure for the GCN, which is orthogonal to this work. 

Non-neural network learning-based approaches have also been applied
to summarization. Typically they involve learning n-gram feature weights 
in linear models along with other non-lexical word or 
structural features 
\cite{berg2011jointly,sipos2012large,durrett2016learning}.
In this paper, we study representation learning in
neural networks that can capture more complex word level feature interactions
and whose dense representations are more compatible with current practices
in NLP.

The previously mentioned works have
focused on news summarization. To further
understand the content selection process, we also explore other domains 
of summarization. In particular, we explore 
personal narrative summarization based on stories shared
on Reddit \cite{ouyang2017crowd}, workplace meeting summarization
\cite{carletta2005ami}, and medical journal article summarization 
\cite{mishra2014text}. 

While most work on these summarization tasks
 often exploit domain-specific features (e.g. speaker identification in meeting summarization \cite{galley2006skip,gillick2009global}),
we purposefully avoid such features in this work in order to understand 
the extent to which deep learning models can perform content 
selection using only surface lexical features.
Summarization of academic literature (including medical journals), has long 
been a research topic in NLP
\cite{kupiec1995trainable,elhadad2005customization}, but most approaches have
explored facet-based summarization \cite{jaidka2017insights}, which is not the focus of our work.

\section{Methods}
\begin{figure*}
  \center
  \includegraphics[scale=.65]{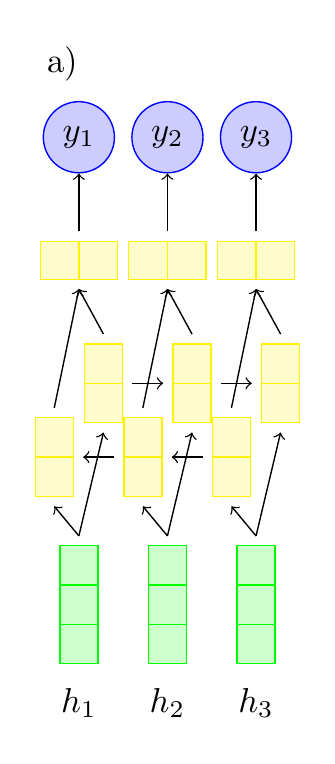}
  \includegraphics[scale=.65]{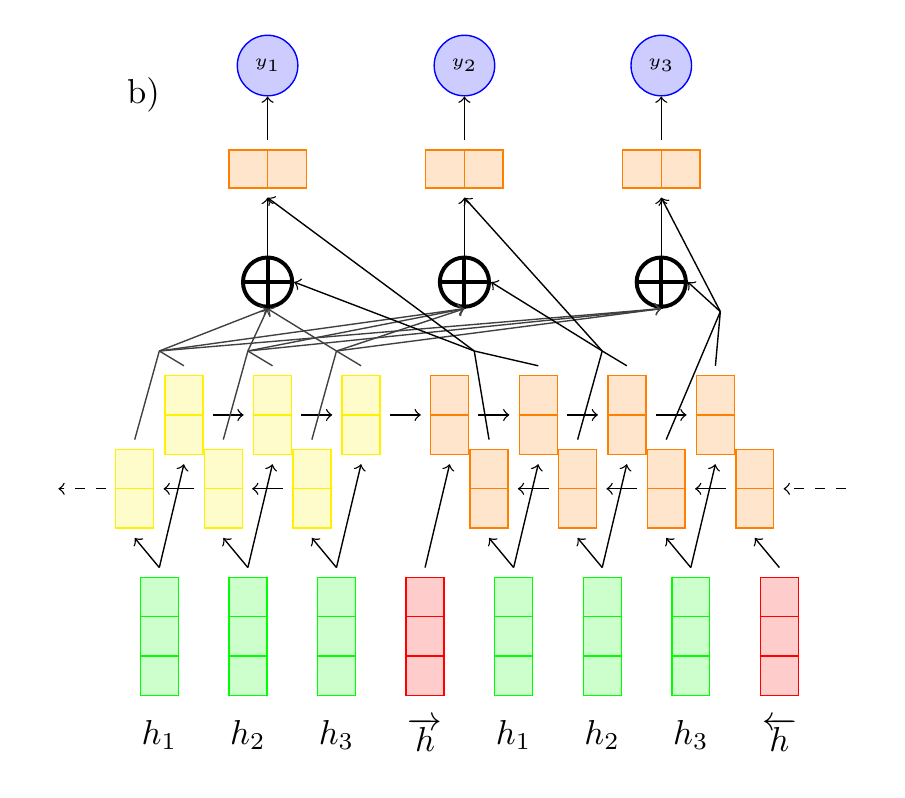}
  \includegraphics[scale=.65]{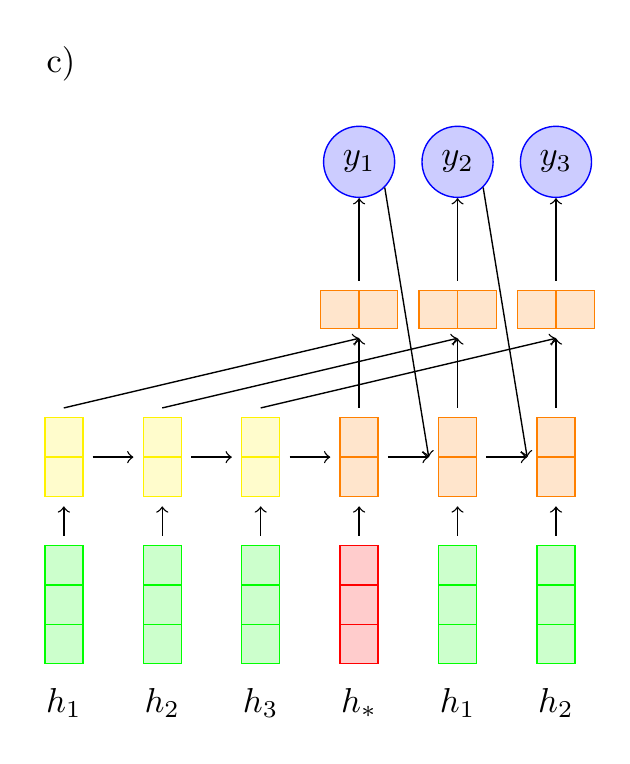}
  \includegraphics[scale=.65]{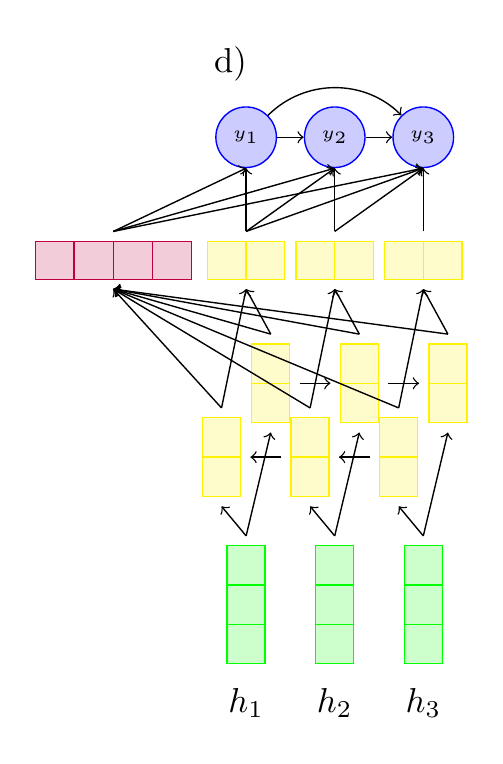}
  \caption{Sentence extractor architectures: a) RNN, b) Seq2Seq,
  c) Cheng \& Lapata, and d) SummaRunner. The $\bigoplus$ indicates 
  attention. Green blocks repesent sentence encoder output and red blocks indicates learned ``begin decoding'' embeddings. Vertically stacked yellow and orange boxes
  indicate
  extractor encoder and decoder hidden states respectively.
  Horizontal orange and yellow blocks indicate multi-layer perceptrons.
  The purple blocks represent the document and summary state in the SummaRunner extractor. }
  \label{fig:extractors}
\end{figure*}

The goal of extractive text summarization is to select a subset of a 
document's text to use as a summary, i.e. a short gist or excerpt of the 
central content.
Typically, we impose a budget on the length of the summary in either 
words or bytes. In this work, we focus on \textit{sentence} extractive 
summarization, 
where the basic unit of extraction is a sentence and impose a word limit as 
the budget.

We model the sentence extraction task as a sequence tagging problem, 
following \cite{conroy2001text}. 
Specifically, given a document containing $\docSize$ sentences 
$\sent_1, \ldots, \sent_{\docSize}$ we generate a summary by predicting a 
corresponding label sequence $\slabel_1, \ldots, \slabel_{\docSize} 
\in \{0, 1\}^{\docSize}$, where $\slabel_i = 1$ 
indicates the $i$-th sentence is to be included in the summary.
Each sentence is itself a sequence of word embeddings 
$\sent[i] = \wordEmb[1]^{(i)}, \ldots, \wordEmb[{|\sent[i]|}]^{(i)}$ where
$|\sent[i]|$ is the length of the sentence in words.
The word budget $c \in \mathbb{N}$ 
enforces a constraint that the total summary word length 
$\sum_{i=1}^\docSize \slabel_i \cdot |\sent[i]| \le c$.


For a typical deep learning model of extractive 
summarization there are two main design decisions:
\textit{a)}  the choice of \textit{sentence encoder} 
which maps each sentence \sent[i] 
to an embedding $\sentEmb[i]$, 
and 
\textit{b)} the choice of \textit{sentence extractor} 
which maps a sequence of sentence embeddings 
$\sentEmb = \sentEmb[1],\ldots, \sentEmb[\docSize]$  
to a sequence of extraction
decisions $\slabel = \slabel_1,\ldots,\slabel_{\docSize}$.

\subsection{Sentence Encoders} \label{sec:senc}
We experiment with three architectures for mapping sequences
of word embeddings to a fixed length vector: averaging, RNNs, and CNNs.
Hyperparameter settings and implementation details can be found 
in \autoref{app:sentencoders}.

\paragraph{Averaging Encoder} Under the averaging encoder, a sentence 
embedding \sentEmb is simply the average of its word embeddings, i.e. 
$\sentEmb = \frac{1}{\sentSize} \sum_{i=1}^{\sentSize} \wordEmb[i]$.

\paragraph{RNN Encoder} When using  the \textit{RNN} sentence encoder,
a sentence embedding is the concatenation of the final output states of a 
forward and backward RNN over the sentence's word embeddings. We use a Gated 
Recurrent Unit (GRU) for the RNN cell \cite{chung2014empirical}.

\paragraph{CNN Encoder} The \textit{CNN} sentence encoder uses a series of 
convolutional feature maps to encode each sentence. This encoder is similar
to the convolutional architecture of \citet{kim2014convolutional} used for 
text classification tasks and performs a series of ``one-dimensional'' 
convolutions over word embeddings. The final sentence embedding $\sentEmb$ is 
a concatenation of all the convolutional filter outputs after max pooling over
time.


\subsection{Sentence Extractors} \label{sec:sext}

Sentence extractors take sentence embeddings $h_{1:n}$ and produce an extract $y_{1:n}$.
The sentence extractor is essentially a discriminative 
classifier $p(\slabel_{1:n} | \sentEmb_{1:n})$.
Previous neural network approaches to sentence extraction have assumed 
an auto-regressive model, leading to a semi-Markovian
factorization of the extractor probabilities 
$p(\slabel_{1:n}|\sentEmb)=\prod_{i=1}^\docSize 
p(\slabel[i]|\slabel[<i],\sentEmb)$,
where each prediction \slabel[i] is dependent on \emph{all} 
previous \slabel[j] for
all $j < i$. We compare two such models proposed by \citet{cheng2016neural}
and \citet{nallapati2017summarunner}.
A simpler approach that does not allow interaction among the $y_{1:n}$
is to 
  model $p(\slabel_{1:n}|\sentEmb) = \prod_{i=1}^n p(y_i|h)$, 
  which we explore in two proposed extractor models that we refer to as the RNN 
  and Seq2Seq extractors.
Implementation details for all extractors are in \autoref{app:sentextractors}.

\paragraph{Previously Proposed Sentence Extractors}
 We consider two recent state-of-the-art extractors.

 The first, proposed by 
\citet{cheng2016neural}, 
is built around a sequence-to-sequence model.
First, each sentence embedding\footnote{\citet{cheng2016neural} used an CNN sentence encoder with 
this extractor architecture; in this work we pair the Cheng \& Lapata extractor
with several different encoders.} is
fed into an encoder side RNN, with the final encoder state passed to the
first step of the decoder RNN. On the decoder side, the same sentence 
embeddings are fed as input to the decoder and decoder outputs are used to
predict each $y_i$. The decoder input is weighted by the previous extraction
probability, inducing the dependence of $y_i$ on $y_{<i}$.
See \autoref{fig:extractors}.c for a graphical layout of the extractor.


\citet{nallapati2017summarunner} proposed
a sentence extractor, which we refer to as the SummaRunner Extractor,
that factorizes the extraction probability into contributions 
from different sources.
First, a bidirectional RNN is run over the sentence embeddings\footnote{\citet{nallapati2017summarunner}
    use an RNN sentence encoder with 
this extractor architecture; in this work we pair the SummaRunner extractor
with different encoders. } and the output is
concatenated. A representation of the whole document is made by 
averaging the RNN output. A summary representation is also constructed 
by taking the sum of the previous RNN outputs weighted by their extraction
probabilities. Extraction predictions are made using 
the RNN output at the $i$-th step, the document representation, and 
$i$-th version of the summary representation, along with factors for 
sentence location in the document. The use of the iteratively constructed
summary representation creates a dependence of $y_i$ on all $y_{<i}$.
See \autoref{fig:extractors}.d for a graphical layout.


\paragraph{Proposed Sentence Extractors}
We propose two sentence extractor models that 
make a stronger conditional independence 
assumption $p(\slabel|\sentEmb)=\prod_{i=1}^\docSize p(\slabel[i]|\sentEmb)$,
essentially making independent predictions conditioned on $\sentEmb$.

\paragraph{RNN Extractor}
    Our first proposed model is a very simple bidirectional
RNN based tagging model. As in the RNN sentence encoder we use a GRU cell.
The forward and backward outputs of each sentence are passed through a 
multi-layer perceptron with a logsitic sigmoid output 
to predict the probability
of extracting each sentence. 
See \autoref{fig:extractors}.a for a graphical layout.

\paragraph{\sts~Extractor} One shortcoming of the RNN extractor is that long range
information from one end of the document may not easily be able to affect 
extraction probabilities of sentences at the other end. 
Our second proposed model, the \sts~extractor mitigates this problem with an 
attention 
mechanism commonly
used for neural machine translation \cite{bahdanau2014neural} and 
abstractive summarization \cite{see2017get}. 
The sentence embeddings are first
encoded by a bidirectional $\gru$. A separate decoder $\gru$ transforms each 
sentence into a query vector which attends to the encoder output. The
attention weighted encoder output and the decoder $\gru$ output are concatenated
and fed into a multi-layer perceptron to compute the extraction probability.
See \autoref{fig:extractors}.b for a graphical layout.

%
%
%
%
%



\begin{table}

    \begin{tabular}{ f  r f r f }
      \toprule
      \textbf{Dataset} & \textbf{Train} & \textbf{Valid} & \textbf{Test} &
        \textbf{Refs} \\
      \midrule
      CNN/DM & 287,113 & 13,368 & 11,490 & 1\\
      NYT & 44,382 & 5,523 & 6,495 & 1.93\\
      DUC & 516 & 91 & 657 & 2 \\
      Reddit & 404 & 24 & 48 & 2 \\
      AMI & 98 & 19 & 20 & 1 \\
      PubMed & 21,250 & 1,250 & 2,500 & 1\\
      \bottomrule
    \end{tabular}
   \caption{Sizes of the training, validation, test splits for each dataset
   and the average number of test set human reference summaries per document.}
   \label{tab:data}
\end{table}

\begin{table*}[ht]
    \center
    \begin{tabular}{ccggccggccggcc}
        \toprule
        \multirow{2}{*}{\textbf{Extractor}} &\multirow{2}{*}{\textbf{Enc.}}  & \multicolumn{2}{g}{\textbf{CNN/DM}} & \multicolumn{2}{c}{\textbf{NYT}} & \multicolumn{2}{g}{\textbf{DUC 2002}} & \multicolumn{2}{c}{\textbf{Reddit}} & \multicolumn{2}{g}{\textbf{AMI}} & \multicolumn{2}{c}{\textbf{PubMed}}\\
         &  & M & R-2 & M & R-2 & M & R-2 & M & R-2 & M & R-2 & M & R-2\\
        \midrule
        Lead &  -- & 24.1 & 24.4 & 30.0 & 32.3 & 25.1 & 21.5 & \textbf{20.1} & \textbf{10.9} & 12.3 &  2.0 & 15.9 &  9.3\\
        \hline
        \multirow{3}{*}{RNN} & Avg. & \textbf{25.2} & 25.4 & 29.8 & 34.7 & \textbf{26.8} & 22.7 & \textbf{20.4} & \textbf{11.4} & \textbf{17.0} & \textbf{ 5.5} & 19.8 & 17.0\\
         & RNN & 25.1 & 25.4 & 29.6 & 34.9 & \textbf{26.8} & 22.6 & \textbf{20.2} & \textbf{11.4} & 16.2 & \textbf{ 5.2} & 19.7 & 16.6\\
         & CNN & 25.0 & 25.1 & 29.0 & 33.7 & \textbf{26.7} & \textbf{22.7} & \textbf{20.9} & \textbf{12.8} & 14.4 &  3.2 & 19.9 & 16.8\\
        \hline
        \multirow{3}{*}{Seq2Seq} & Avg. & \textbf{25.2} & \textbf{25.6} & \textbf{30.5} & \textbf{35.7} & \textbf{27.0} & \textbf{22.8} & \textbf{20.9} & \textbf{13.6} & \textbf{17.0} & \textbf{ 5.5} & \textbf{20.1} & \textbf{17.7}\\
         & RNN & \textbf{25.1} & 25.3 & 30.2 & \textbf{35.9} & \textbf{26.7} & 22.5 & \textbf{20.5} & \textbf{12.0} & 16.1 & \textbf{ 5.3} & 19.7 & 16.7\\
         & CNN & 25.0 & 25.1 & 29.9 & 35.1 & \textbf{26.7} & \textbf{22.7} & \textbf{20.7} & \textbf{13.2} & 14.2 &  2.9 & 19.8 & 16.9\\
        \hline
    \multirow{3}{*}{\begin{tabular}{c} Cheng \\ \& \\ Lapata \end{tabular}} & Avg. & 25.0 & 25.3 & 30.4 & \textbf{35.6} & \textbf{27.1} & \textbf{23.1} & \textbf{20.9} & \textbf{13.6} & \textbf{16.7} & \textbf{ 6.1} & \textbf{20.1} & \textbf{17.7}\\
         & RNN & 25.0 & 25.0 & \textbf{30.3} & \textbf{35.8} & \textbf{27.0} & \textbf{23.0} & \textbf{20.3} & \textbf{12.6} & \textbf{16.3} & \textbf{ 5.0} & 19.7 & 16.7\\
         & CNN & \textbf{25.2} & 25.1 & 29.9 & 35.0 & \textbf{26.9} & \textbf{23.0} & \textbf{20.5} & \textbf{13.4} & 14.3 &  2.8 & 19.9 & 16.9\\
        \hline
    \multirow{3}{*}{\begin{tabular}{c}Summa \\ Runner \end{tabular} } & Avg. & 25.1 & 25.4 & 30.2 & 35.4 & 26.7 & 22.3 & \textbf{21.0} & \textbf{13.4} & \textbf{17.0} & \textbf{ 5.6} & 19.9 & 17.2\\
         & RNN & 25.1 & 25.2 & 30.0 & 35.5 & 26.5 & 22.1 & \textbf{20.9} & \textbf{12.5} & \textbf{16.5} & \textbf{ 5.4} & 19.7 & 16.5\\
         & CNN & 24.9 & 25.0 & 29.3 & 34.4 & 26.4 & 22.2 & \textbf{20.4} & \textbf{12.3} & 14.5 &  3.2 & 19.8 & 16.8\\
        \hline
        Oracle & -- & 31.1 & 36.2 & 35.3 & 48.9 & 31.3 &  31.8 & 24.3 &  16.2 &17.8 &  8.7  & 24.1 &25.0 \\
        \bottomrule
    \end{tabular}

    \caption{METEOR (M) and ROUGE-2 recall (R-2)  results across all 
        extractor/encoder pairs.
           Results that are statistically indistinguishable from the best 
           system are shown in bold face.}
  \label{tab:results}
\end{table*}

\section{Datasets}
\label{sec:datasets}
We perform our experiments across six corpora from varying domains to 
understand how different biases within each domain can affect content 
selection. The corpora come from the news domain
(CNN-DailyMail, New York Times, DUC), personal narratives domain (Reddit),
workplace meetings (AMI), and medical journal articles (PubMed). See 
\autoref{tab:data} for dataset statistics.

\paragraph{CNN-DailyMail} We use the preprocessing and training, validation, 
and test splits
of \citet{see2017get}.
This corpus is a mix of news on different topics including politics,
sports, and entertainment.

\paragraph{New York Times}The New York Times (NYT) corpus \cite{sandhaus2008new} contains
 two types of abstracts for a subset of its articles. The first summary is
an archival abstract and the 
second is a shorter online teaser meant to entice a viewer of the webpage to
click to read more. From this collection, we take all articles that have 
a concatenated summary length of at least 100 words.
We create training, validation, and test splits by partitioning on dates;
we use the year 2005 as the validation data, with training and test partitions
including documents before and after 2005 respectively.

\paragraph{DUC} We use the single document summarization data from the 2001
and 2002
Document Understanding Conferences (DUC) \cite{over2002introduction}. We split the 2001 data into training
and validation splits and reserve the 2002 data for testing.

\paragraph{AMI} The AMI corpus \cite{carletta2005ami} 
is a collection of real and staged office meetings
annotated with text transcriptions, along with abstractive
summaries. We use the prescribed splits. 

\paragraph{Reddit} \citet{ouyang2017crowd} collected a corpus of personal 
    stories shared
 on Reddit\footnote{\url{www.reddit.com}} along with multiple extractive 
 and abstractive summaries. We randomly split this data using roughly three and five percent of the data validation and test respectively.

\paragraph{PubMed}{We created a corpus of 25,000 randomly sampled
    medical journal articles from the PubMed Open Access 
    Subset\footnote{\url{https://www.ncbi.nlm.nih.gov/pmc/tools/openftlist/}}.
    We only included articles if they were at least 1000 words long and 
    had an abstract of at least 50 words in length.
We used the article abstracts as the ground truth human summaries.}

\subsection{Ground Truth Extract Summaries}
Since we do not typically have ground truth extract summaries from which to
create the labels $\slabel_i$, we construct gold label sequences 
by greedily optimizing ROUGE-1, using the algorithm in \autoref{app:oracle}.
We choose to optimize for ROUGE-1 rather than 
ROUGE-2 similarly to other optimization based approaches to summarization 
\cite{sipos2012large,durrett2016learning} which found this to
be the easier target to learn.


\begin{table*}[t]
\center
\begin{tabular}{ccgL{.5cm}cm{.5cm}gL{.75cm}cm{.75cm}gL{.75cm}cm{.5cm}}
    \toprule
    \multirow{1}{*}{\textbf{Ext.}} &\multirow{1}{*}{\textbf{Emb.}}  & \multicolumn{2}{g}{\textbf{CNN/DM}} & \multicolumn{2}{c}{\textbf{NYT}} & \multicolumn{2}{g}{\textbf{DUC}} & \multicolumn{2}{c}{\textbf{Reddit}} & \multicolumn{2}{g}{\textbf{AMI}} & \multicolumn{2}{c}{\textbf{PubMed}}\\
    \midrule
    \multirow{2}{*}{Seq2Seq} & Fixed & \textbf{25.6} && \textbf{35.7}& & \textbf{22.8}& & \textbf{13.6} &&  5.5 && \textbf{17.7}\\
   & Learn & 25.3 &\footnotesize{(0.3)} & \textbf{35.7}& \footnotesize{(0.0)} & \textbf{22.9} & \footnotesize{(-0.1)} & \textbf{13.8} &\footnotesize{ (-0.2)} & \textbf{ 5.8} & \footnotesize{(-0.3)} & 16.9 & \footnotesize{(0.8)}\\
    \hline
    \multirow{2}{*}{C\&L} & Fixed & \textbf{25.3} && \textbf{35.6} && \textbf{23.1} && \textbf{13.6} && \textbf{ 6.1} && \textbf{17.7}&\\
                      & Learn & 24.9 &\footnotesize{(0.4)} & 35.4 & \footnotesize{(0.2)} & \textbf{23.0} &\footnotesize{ (0.1)} & \textbf{13.4} &\footnotesize{ (0.2)} & \textbf{ 6.2} &\footnotesize{ (-0.1)} & 16.4 &\footnotesize{ (1.3)} \\
    \hline
\multirow{2}{*}{\begin{tabular}{c} Summa \\ Runner \end{tabular}} & Fixed & \textbf{25.4} && \textbf{35.4} && \textbf{22.3} && \textbf{13.4} && \textbf{ 5.6} & &\textbf{17.2}&\\
                                                                  & Learn & 25.1 &\footnotesize{(0.3)} & 35.2 &\footnotesize{(0.2)} & \textbf{22.2} & \footnotesize{(0.1)} & 12.6 & \footnotesize{(0.8)} & \textbf{ 5.8} & \footnotesize{(-0.2)} & 16.8&\footnotesize{ (0.4) }\\
    \bottomrule
\end{tabular}

\caption{ROUGE-2 recall across sentence extractors
    when using fixed pretrained embeddings or when embeddings are updated during training. In both cases embeddings
    are initialized with pretrained GloVe embeddings. All extractors use the averaging 
sentence encoder. When both learned and fixed settings are bolded,
there is no signifcant performance difference. RNN extractor is omitted for space but is similar to Seq2Seq. Difference in scores shown in parenthesis.}
\label{tab:embeddings}
\end{table*}


\begin{table*}[ht]
\center
\begin{tabular}{cgcgcgc}
    \toprule
    \multirow{1}{*}{\textbf{Ablation}}  & \multicolumn{1}{g}{\textbf{CNN/DM}} & \multicolumn{1}{c}{\textbf{NYT}} & \multicolumn{1}{g}{\textbf{DUC}} & \multicolumn{1}{c}{\textbf{Reddit}} & \multicolumn{1}{g}{\textbf{AMI}} & \multicolumn{1}{c}{\textbf{PubMed}}\\
    \hline
    all words & \textbf{25.4}\textsuperscript{~} ~~~~~~~ & \textbf{34.7}\textsuperscript{~} ~~~~~~~~& 22.7\textsuperscript{~} ~~~~~~~~& \textbf{11.4}\textsuperscript{~} ~~~~~~~~& 5.5\textsuperscript{~} ~~~~~~~~~& \textbf{17.0}\textsuperscript{~} ~~~~~~~ \\
    -nouns & 25.3\textsuperscript{$\dagger$} \footnotesize{(0.1)}& 34.3\textsuperscript{$\dagger$} \footnotesize{(0.4)}& 22.3\textsuperscript{$\dagger$} ~\footnotesize{(0.4)}& 10.3\textsuperscript{$\dagger$} \footnotesize{(1.1)} & 3.8\textsuperscript{$\dagger$} \footnotesize{(1.7)}& 15.7\textsuperscript{$\dagger$} \footnotesize{(1.3)}\\
    -verbs & 25.3\textsuperscript{$\dagger$} \footnotesize{(0.1)}& 34.4\textsuperscript{$\dagger$} \footnotesize{(0.3)} & 22.4\textsuperscript{$\dagger$} ~\footnotesize{(0.3)}& 10.8\textsuperscript{~} ~\footnotesize{(0.6)} & 5.8\textsuperscript{~} \footnotesize{(-0.3)} & 16.6\textsuperscript{$\dagger$} \footnotesize{(0.4)}\\
    -adj/adv & 25.3\textsuperscript{$\dagger$} \footnotesize{(0.1)}& 34.4\textsuperscript{$\dagger$} \footnotesize{(0.3)} & 22.5\textsuperscript{~} ~\footnotesize{(0.2)} & ~~9.5\textsuperscript{$\dagger$} \footnotesize{(1.9)} & 5.4\textsuperscript{~} ~\footnotesize{(0.1)} & 16.8\textsuperscript{$\dagger$} \footnotesize{(0.2)}\\
    -function & 25.2\textsuperscript{$\dagger$} \footnotesize{(0.2)} & 34.5\textsuperscript{$\dagger$} \footnotesize{(0.2)} & \textbf{22.9}\textsuperscript{$\dagger$} \footnotesize{(-0.2)} & 10.3\textsuperscript{$\dagger$} \footnotesize{(1.1)}& \textbf{6.3}\textsuperscript{$\dagger$} \footnotesize{(-0.8)}& 16.6\textsuperscript{$\dagger$} \footnotesize{(0.4)}\\
    \bottomrule
\end{tabular}

\caption{ROUGE-2 recall after removing nouns, verbs, adjectives/adverbs, and 
    function words. Ablations are
    performed using the averaging sentence encoder and the RNN
extractor. 
Bold indicates best performing system. $\dagger$ indicates significant 
difference with the non-ablated system. Difference in score from \textit{all words} shown in parenthesis.}
\label{tab:ablations}
\end{table*}

\begin{table*}[ht]
\center

\begin{tabular}{ccgL{.5cm}cm{.5cm}gL{.5cm}cm{.75cm}gL{.75cm}cm{.5cm}}
    \toprule
    \textbf{Ext.} &\textbf{Order}  & \multicolumn{2}{g}{\textbf{CNN/DM}} & \multicolumn{2}{c}{\textbf{NYT}} & \multicolumn{2}{g}{\textbf{DUC}} & \multicolumn{2}{c}{\textbf{Reddit}} & \multicolumn{2}{g}{\textbf{AMI}} & \multicolumn{2}{c}{\textbf{PubMed}}\\
    \midrule
    \multirow{2}{*}{Seq2Seq} & In-Order & \textbf{25.6} & & \textbf{35.7} && \textbf{22.8}& & \textbf{13.6} && 5.5 && \textbf{17.7} &\\
                             & Shuffled & 21.7&\footnotesize{(3.9)} & 25.6 & \footnotesize{(10.1)} & 21.2 & \footnotesize{(1.6)} &\textbf{13.5} &\footnotesize{(0.1)} &\textbf{6.0} & \footnotesize{(-0.5)}&14.9 &\footnotesize{(2.8)}\\
    \bottomrule
\end{tabular}

\caption{ROUGE-2 recall using models trained on in-order and shuffled
documents. Extractor uses the averaging sentence encoder. 
When both in-order and shuffled settings are bolded,
there is no signifcant performance difference. Difference in scores shown in parenthesis.
}
\label{tab:shuffle}
\end{table*}

\section{Experiments} \label{sec:exps}

We evaluate summary quality using ROUGE-2 recall \cite{lin2004rouge};
ROUGE-1 and ROUGE-LCS trend similarity in our experiments.
We use target word lengths of 100 words for news, and 
75, 290, and 200 for Reddit, AMI, and PubMed respectively.
We also evaluate using METEOR \cite{denkowski:lavie:meteor-wmt:2014}.\footnote{We use the default settings for METEOR and use remove stopwords and no stemming options for ROUGE, keeping defaults for all other parameters.}
Summaries are generated by extracting the top ranked sentences by model probability $p(y_i=1|y_{<i},h)$, stopping when the word budget is met or exceeded.
We estimate statistical significance by averaging each document level score
over the five random initializations. 
We then test the difference between the best system on each dataset and 
all other systems using the approximate randomization test 
\cite{riezler2005some} with the Bonferroni correction for multiple comparisons,
testing for significance at the $0.05$ level. 

\subsection{Training}

We train all models to minimize the weighted negative log-likelihood
\[\mathcal{L} = -\sum_{\substack{ s,y\in \mathcal{D} \\ h = \operatorname{enc}(s) } } \sum_{i=1}^\docSize \omega(y_i) \log p\left(y_i|y_{<i},
h \right)\]
over the training data $\mathcal{D}$
using stochastic gradient descent with the ADAM optimizer
\cite{kingma2014adam}.
$\omega(0)=1$ and $\omega(1) = N_0/N_1$ where $N_y$ is the number of 
training examples with label $y$.
    We trained for a maximum of 50 epochs and the best
    model was selected with early stopping on the validation set according
    to ROUGE-2. Each epoch constitutes a full pass through the
    dataset. The average stopping epoch was: CNN-DailyMail, 16.2; NYT, 21.36; DUC, 37.11; Reddit, 36.59; AMI, 19.58; PubMed, 19.84.
     All experiments were repeated with five random
    initializations.     Unless specified, word embeddings were initialized 
    using pretrained GloVe embeddings \cite{pennington2014glove} and we did 
    not update them during training. Unknown words were mapped to a zero 
    embedding.
    See \autoref{app:optset} for more optimization and training details.

\subsection{Baselines}
\paragraph{Lead} As a baseline we include the lead summary, i.e. taking the first 
$x$ words of the document as summary, where $x$ is the target summary length for each dataset (see the 
first paragraph of \S~\ref{sec:exps}). While incredibly simple, this method is still a 
competitive baseline for single document summarization, especially on newswire.
\paragraph{Oracle} To measure the performance ceiling,
we show the ROUGE/METEOR scores using the 
extractive summary which results from greedily optimizing ROUGE-1. I.e., if we 
had clairvoyant knowledge
of the human reference summary, the oracle system achieves the (approximate) 
maximum possible ROUGE scores. 
See \autoref{app:oracle} for a detailed
description of the oracle algorithm.

\begin{table*}[ht]
    \footnotesize
\centering
  \begin{tabular}{p{24em} p{24em}}
\toprule
Hurricane Gilbert swept toward the Dominican Republic Sunday, and the 
   Civil Defense alerted its heavily populated south coast to prepare 
   for high winds, heavy rains and high seas.                         
The storm was approaching from the southeast with sustained winds of  
   75 mph gusting to 92 mph.                                          
An estimated 100,000 people live in the province, including 70,000 in 
   the city of Barahona, about 125 miles west of Santo Domingo.       
\textbf{On Saturday, Hurricane Florence was downgraded to a tropical storm and
   its remnants pushed inland from the U.S. Gulf Coast.}               
Tropical Storm Gilbert formed in the eastern Caribbean and            
   strengthened into a hurricane Saturday night.  
&
Hurricane Gilbert swept toward the Dominican Republic Sunday, and the 
   Civil Defense alerted its heavily populated south coast to prepare 
   for high winds, heavy rains and high seas.                         
The storm was approaching from the southeast with sustained winds of  
   75 mph gusting to 92 mph.                                          
An estimated 100,000 people live in the province, including 70,000 in 
   the city of Barahona, about 125 miles west of Santo Domingo.       
Tropical Storm Gilbert formed in the eastern Caribbean and            
   strengthened into a hurricane Saturday night.                      
\textbf{Strong winds associated with the Gilbert brought coastal flooding,    
   strong southeast winds and up to 12 feet feet to Puerto Rico's     
   south coast.}   \\
\bottomrule
\end{tabular}
\caption{Example output of Seq2Seq extractor (left) and Cheng 
\& Lapata Extractor (right). This is a typical example, where only one
 sentence is different between the two (shown in bold). }
\label{tab:output}
\end{table*}

\subsection{Results}

The results of our main experiment comparing 
the different extractors/encoders are shown in 
Table~\ref{tab:results}.
Overall, we find no major advantage when using the CNN and RNN sentence
encoders over the averaging encoder. The best performing encoder/extractor pair either 
uses the averaging 
encoder (five out of six datasets) or the differences 
are not statistically significant. 


When looking at extractors, the Seq2Seq extractor is either part of 
the best performing system (three out of six datasets) or is not 
statistically distinguishable from the best extractor. 

Overall, on the news and medical journal domains, the differences are 
quite small with the 
differences between worst and best systems on the CNN/DM dataset 
spanning only .56 of a ROUGE point. While there is more performance variability
 in the Reddit and AMI data, there is less distinction among systems: 
 no differences are significant on Reddit
and every extractor has at least one configuration that is indistinguishable
from the best system on the AMI corpus. This is probably due to the small test
size of these datasets.



\paragraph{Word Embedding Learning}
 Given that learning a sentence encoder (averaging has no learned parameters)
 does not yield significant improvement, it is natural to consider whether
 learning word embeddings is also necessary. 
 In \autoref{tab:embeddings} we compare the performance of different extractors
 using the averaging encoder, when the word embeddings are held fixed or 
 learned during training. In both cases, word embeddings are initialized with
 GloVe embeddings trained on a combination of Gigaword and Wikipedia.
 When learning embeddings, words occurring 
 fewer than three times in the training data are mapped to an unknown
 token (with learned embedding).
 
 In all but one case,
fixed embeddings are as good or better than the learned embeddings.
This is a somewhat surprising finding on the CNN/DM data since it is reasonably
large, and learning embeddings should give the models more
flexibility to identify important word features.\footnote{The AMI corpus is an exception here where learning \emph{does} lead to small
performance boosts, however, only in the Seq2Seq extractor is this diference 
significant; it is quite possible that this is an artifact of the very small
test set size.}
This suggests that we cannot extract much generalizable learning signal 
from the content other than what is already present from initialization. 
Even on PubMed, where the language is quite different from the news/Wikipedia
articles the GloVe embeddings were trained on, learning leads to 
significantly worse results.



\paragraph{POS Tag Ablation}
It is also not well explored what word features are being used by the encoders.
To understand which classes of words were most important we ran an ablation
study, selectively removing nouns, verbs 
(including participles and auxiliaries), adjectives \& adverbs, and 
function words (adpositions, determiners, conjunctions).
All datasets were automatically tagged using
the spaCy part-of-speech (POS)
tagger\footnote{https://github.com/explosion/spaCy}.   
The embeddings of removed words were replaced with a zero vector,
preserving the order and position of the non-ablated words in the sentence.
Ablations were performed on training, validation, and test partitions,
using the RNN extractor with averaging encoder.
\autoref{tab:ablations} shows the results of the POS
tag ablation experiments. 
While removing any word class from the representation generally hurts 
performance (with statistical significance), on the news domains,
the absolute values of the differences are quite small 
(.18 on CNN/DM, .41 on NYT, .3 on DUC) suggesting that the model's predictions
are not overly dependent on any particular word types.
On the non-news datasets, the ablations have a larger effect 
(max differences are 1.89 on Reddit, 2.56 on AMI, and 1.3 on PubMed).
Removing nouns leads to the largest drop on AMI and PubMed.
Removing adjectives and adverbs leads to the largest drop on Reddit,
suggesting the intensifiers and descriptive words are useful for 
identifying important content in personal narratives.
Curiously, 
removing the function word POS class yields a significant improvement
on DUC 2002 and AMI.


\textbf{Document Shuffling} Sentence position is a well known and 
powerful feature for news summarization \cite{hong2014improving}, owing 
to the intentional lead bias in the news article writing\footnote{\url{https://en.wikipedia.org/wiki/Inverted_pyramid_(journalism)}}; it also explains the difficulty in beating
the lead baseline for single-document summarization 
\cite{nenkova2005automatic,rau:1999}.
In examining the generated summaries, we found
most of the selected sentences in the news domain came from the lead paragraph
of the document. This is despite the fact that there is a long tail of 
sentence extractions from later in the document in the ground truth extract 
summaries (31\%, 28.3\%, and 11.4\% of DUC, CNN/DM, and NYT training extract labels come 
from the second half of the document). 
Because this lead bias is so strong, it is questionable whether
the models are learning to identify important content or just find the start
of the document. We conduct a sentence order experiment where 
each document's sentences are randomly shuffled during training. We then
evaluate each model performance on the unshuffled test data, comparing to 
the model trained on unshuffled data; if the models trained on shuffled data
drop in performance, then this indicates the lead bias is the relevant factor.

\autoref{tab:shuffle} shows the results
of the shuffling experiments. 
The news domains and PubMed suffer a significant drop in performance 
when the document order is shuffled. By comparison, there is no significant difference between the shuffled and in-order models on 
the Reddit domain, and shuffling actually improves performance on AMI.
This suggest that position 
is being learned by the models in the news/journal article domain even when 
the model has no explicit position features, and that this feature is more 
important than either content or function words.



\section{Discussion}

\hal{MAKE SURE THE MAIN POINTS FROM INTRO ARE ALLUDED TO SOMEWHERE}
Learning content selection for summarization in the news domain is severely inhibited by the lead bias. 
The summaries generated by all systems described here--the prior work and our proposed simplified models--are highly similar to each other and to the lead 
baseline. The Cheng \& Lapata and Seq2Seq 
extractors (using the averaging encoder) share 87.8\% of output sentences on average on the CNN/DM data,
with similar numbers for the other news domains (see \autoref{tab:output}
for a typical example).  
Also on CNN/DM, 58\% of the Seq2Seq 
selected sentences also occur
in the lead summary, with similar numbers for DUC, NYT, and Reddit. Shuffling
reduces lead overlap to 35.2\% but the overall system performance drops
    significantly; the models are not able to identify important information
    without position.
    
    The relative robustness of the news domain to part of speech ablation also 
    suggests that models are mostly learning to recognize the stylistic 
    features unique to the beginning of the article, and not the content.
    Additionally, the drop in performance when learning word embeddings on 
    the news domain suggests that word embeddings alone do not provide 
    very generalizable content features compared to recognizing the lead.

The picture is rosier for non-news summarization where part of speech ablation leads
to larger performance differences and shuffling either does not inhibit content
selection significantly or leads to modest gains. Learning better
word-level representations on these domains will likely require much
larger corpora, something which might remain unlikely for personal stories
and meetings.

The lack of distinction among sentence encoders is interesting because 
it echoes findings in the generic sentence embedding literature 
where word embedding averaging is frustratingly difficult to 
outperform  \cite{iyyer2015deep,wieting2015towards,arora2016simple,wieting2017revisiting}.
The inability to learn useful sentence representations is also 
borne out in the 
SummaRunner model, where there are explicit similarity computations
between document or summary representations and sentence embeddings;
these computations do not seem to add much to the performance as the 
Cheng \& Lapata and Seq2Seq models which lack these features generally
perform as well or better.
Furthermore, the Cheng \& Lapata and SummaRunner extractors both construct
a history of previous selection decisions to inform future choices but this
does not seem to significantly improve performance over the Seq2Seq extractor 
(which does not). This suggests that we need to rethink or find novel forms 
of sentence representation for the summarization task.\hal{I'M NOT SURE ABOUT THAT. ISN'T IT MORE SAYING THAT THE INPUT REPRESENTATION IS SUFFICIENTLY RICH THAT DEPENDENCIES IN THE OUTPUT DO NOT NEED TO BE MODELED EXPLICITLY?}

A manual examination of the outputs revealed some interesting failure modes,
although in general it was hard to discern clear patterns of behaviour 
other than lead bias. On the news domain, the models consistently learned 
to ignore quoted material in the lead, as often the quotes provide
color to the story but are unlikely to be included in the summary (e.g. \textit{``It was like somebody slugging a punching bag.''}). 
This behavior was most likely triggered by the presence of quotes, as the
quote attributions, which were often tokenized as separate sentences,
would subsequently be included in the summary despite also not containing 
much information 
(e.g. \textit{Gil Clark of the National Hurricane Center said Thursday}). \hal{DOES THE POS ABLATION HAVE ANYTHING TO SAY HERE?}

%

\section{Conclusion}
We have presented an empirical study of deep learning based content selection
algorithms for summarization. Our findings suggest such models face stark limitations on their ability to learn robust features for this task and that 
more work is needed on sentence representation for summarization.

\section{Acknowledgements}
The authors would like to thank the anonymous reviewers for their valuable 
feedback. Thanks goes out as well to Chris Hidey for his helpful comments. We would also like to thank Wen Xiao for identifying an error in the oracle 
results for the AMI corpus, which as since been corrected.

This research is based upon work supported in part by the Office of the 
Director of National Intelligence (ODNI), Intelligence Advanced Research 
Projects Activity (IARPA), via contract \# FA8650-17-C-9117. The views and 
conclusions contained herein are those of the authors and should not be 
interpreted as necessarily representing the official policies, either 
expressed or implied, of ODNI, IARPA, or the U.S. Government. The U.S. 
Government is authorized to reproduce and distribute reprints for governmental 
purposes notwithstanding any copyright annotation therein.

\bibliography{emnlp2018}
\bibliographystyle{acl_natbib_nourl}

\newpage
\onecolumn
\appendix

\begin{centering}
  \Large
  Supplementary Material For:\\
  Content Selection in Deep Learning Models of Summarization\\

\end{centering}

  \section{Details on Sentence Encoders} \label{app:sentencoders}

  We use 200 dimenional word embeddings \wordEmb[i] in all models.
  Dropout is applied to the embeddings during training. 
  Wherever dropout is applied, the drop probability is .25.

\subsection{Details on RNN Encoder} \label{app:rnnencoder}
  
Under the \textit{RNN} encoder, a sentence embedding is defined as
$\sentEmb = [\rSentEmb[\sentSize]; \lSentEmb[1]]$
where 
\begin{align} 
  \rSentEmb[0] = \mathbf{0} &;\quad 
       \rSentEmb[i] = \rgru(\wordEmb[i], \rSentEmb[i-1]) \\
  \lSentEmb[\sentSize + 1] = \mathbf{0} &;\quad 
       \lSentEmb[i] = \lgru(\wordEmb[i], \lSentEmb[i+1]),
\end{align}
and $\rgru$ amd $\lgru$ indicate the 
forward and backward GRUs respectively, each with separate 
parameters. We use 300 dimensional hidden layers for each GRU. 
Dropout is applied to GRU during training.

\subsection{Details on CNN Encoder} \label{app:cnnencoder}
The \textit{CNN} encoder has hyperparameters
associated with the window sizes $\maxWindowSize \subset \mathbb{N}$ of the convolutional filters
(i.e. the number of words associated with each convolution) and the number of 
feature maps $\maxFeatureMaps_k \in \mathbb{N}$ associated with each filter
(i.e. the output dimension of each 
convolution). 
The \textit{CNN} sentence embedding $\sentEmb$ is computed as follows:
\begin{align}
 \specActivation_i &= \specConvBias 
    + \sum^\filterWindowSize_{j=1} \specConvWeight_j \cdot \wemb_{i + j -1}\\
  \specFeatureMap &= \max_{i\in 1,\dots, |\sent| - \filterWindowSize + 1} 
                      \relu\left(\specActivation_i\right) \\
 \sentEmb &= \left[\specFeatureMap | 
   \numFeatureMaps \in \{1, \ldots, \maxFeatureMaps_k\},
   \filterWindowSize \in \maxWindowSize
   \right]
\end{align}
where $\specConvBias\in\mathcal{R}$ and $\specConvWeight \in 
\mathcal{R}^{\filterWindowSize \times \wordEmbSize}$ are learned bias and filter
weight parameters respectively, and $\relu(x) = \max(0, x)$ is the rectified
linear unit activation.
We use window sizes $K=\{1, 2, 3, 4, 5, 6\}$ with corresponding feature maps sizes $M_1=25, M_2=25, M_3=50, M_4=50, M_5=50, M_6=50$, giving $h$ a dimensionality of 250. 
Dropout is applied to the CNN output during training.

\section{Details on Sentence Extractors} \label{app:sentextractors}
\subsection{Details on RNN Extractor} \label{app:rnnextractor}
\begin{align}
    \rExtHidden_0 = \mathbf{0};&\quad   \rExtHidden_i = \rgru(\sentEmb[i], \rExtHidden_{i-1}) \\
    \lExtHidden_{\docSize + 1} = \mathbf{0};&\quad    \lExtHidden_i = \lgru(\sentEmb[i], \lExtHidden_{i+1}) \\
   \logits_i &= \relu\left(U \cdot [\rExtHidden_i; \lExtHidden_i] + u \right)\\
   p(\slabel_i=1|\sentvec) &= \sigma\left(V\cdot \logits_i + v  \right)
\end{align}
where $\rgru$ and $\lgru$ indicate the 
forward and backward GRUs respectively, and each have separate learned 
parameters; $U, V$ and $u, v$ are learned weight and bias parameters.
The hidden layer size of the GRU is 300 for each direction and the MLP hidden layer
size is 100. Dropout is applied to the GRUs and to $a_i$.

\subsection{Details on Seq2Seq Extrator} \label{app:s2sextractor}
\begin{align}
    \rEncExtHidden_0 = \textbf{0}&;\quad \rEncExtHidden_i = \rgru_{enc}(\sentEmb[i], \rEncExtHidden_{i-1}) \\
    \lEncExtHidden_{\docSize + 1} = \textbf{0}&;\quad  \lEncExtHidden_i = \lgru_{enc}(\sentEmb[i], \lEncExtHidden_{i+1}) \\
    \rDecExtHidden_i &= \rgru_{dec}(\sentEmb[i], \rDecExtHidden_{i-1}) \\
    \lDecExtHidden_i &= \lgru_{dec}(\sentEmb[i], \lDecExtHidden_{i+1}) 
\end{align}
\begin{align}
 \decExtHidden_i = [\rDecExtHidden_i; \lDecExtHidden_i], &\;\;
 \encExtHidden_i = [\rEncExtHidden_i; \lEncExtHidden_i] 
\end{align}
\begin{align}
 \alpha_{i,j} = 
   \frac{\exp \left(\decExtHidden_i \cdot \encExtHidden_j \right)}{
   \sum_{j=1}^{\docSize}\exp\left(\decExtHidden_i \cdot \encExtHidden_j\right)}, 
& \;\; \attnExtHidden_i = \sum_{j=1}^{\docSize} \alpha_{i,j} \encExtHidden_j 
\end{align}
\begin{align}
   \logits_i = \relu\left(U \cdot [\attnExtHidden_i; \decExtHidden_i] + u \right)&\\
   p(\slabel_i=1|\sentvec) = \sigma\left(V\cdot \logits_i + v  \right).
\end{align}
The final outputs of each encoder direction are passed to the first decoder
steps; additionally, the first step of the decoder GRUs are learned 
``begin decoding'' vectors $\rDecExtHidden_0$ and $\lDecExtHidden_0$ 
(see \autoref{fig:extractors}.b).
Each GRU has separate learned 
parameters; $U, V$ and $u, v$ are learned weight and bias parameters.
The hidden layer size of the GRU is 300 for each direction and MLP hidden layer
size is 100. Dropout with drop probability .25 is applied to the GRU outputs and to $a_i$.

\subsection{Details on Cheng \& Lapata Extractor.} \label{app:clextractor}
The basic architecture is a unidirectional
sequence-to-sequence
model defined as follows:
\begin{align}
    \encExtHidden_0 = \textbf{0};&\quad   \encExtHidden_i = \gru_{enc}(\sentvec_i, \encExtHidden_{i-1}) \\
    \decExtHidden_1 &= \gru_{dec}(\sentEmb[*], \encExtHidden_{\docSize}) \\
    \decExtHidden_i &= \gru_{dec}(p_{i-1} \cdot \sentvec_{i-1}, \decExtHidden_{i-1}) \label{eq:cl1} \\
   \logits_i &= \relu\left(U \cdot [\encExtHidden_i; \decExtHidden_i] + u \right)\\
    p_i = p(\slabel_i&=1|\slabel_{<i}, \sentvec) = \sigma\left(V\cdot \logits_i + v  \right) 
\end{align}
where \sentEmb[*] is a learned ``begin decoding'' sentence embedding
(see \autoref{fig:extractors}.c).
Each GRU has separate learned 
parameters; $U, V$ and $u, v$ are learned weight and bias parameters.
Note in Equation~\ref{eq:cl1} that 
the decoder side GRU input is the sentence embedding from the previous time
step weighted by its probabilitiy of extraction ($p_{i-1}$) from the 
previous step, inducing dependence of each output $y_i$ on all previous 
outputs $y_{<i}$.
The hidden layer size of the GRU is 300 and the MLP hidden layer
size is 100. 
Dropout with drop probability .25 is applied to the GRU outputs and to $a_i$.


Note that in the original paper, the Cheng \& Lapata extractor was paired 
with
a \textit{CNN} sentence encoder, but in this work we experiment with a variety
of sentence encoders.

\subsection{Details on SummaRunner Extractor.} \label{app:srextractor}
Like the
RNN~extractor it starts with a bidrectional GRU over the sentence 
embeddings 
\begin{align}
    \rEncExtHidden_0 = \textbf{0}&;\quad \rEncExtHidden_i = \rgru(\sentvec_i, \rEncExtHidden_{i-1}) \\
    \lEncExtHidden_{\docSize + 1} = \textbf{0}&;\quad \lEncExtHidden_i = \lgru(\sentvec_i, \lEncExtHidden_{i+1}),
\end{align}

It then creates a representation
of the whole document $q$ by passing the averaged GRU output states through
a fully connected layer: 
\begin{align}
q = \tanh\left(b_q + W_q\frac{1}{\docSize}\sum_{i=1}^{\docSize} [\rEncExtHidden_i; \lEncExtHidden_i] \right)
\end{align}
A concatentation of the GRU outputs at each step
are passed through a separate fully connected layer to create a 
sentence representation $z_i$, where
\begin{align}
    \extHidden_i &= \relu\left(b_z + W_z [\rEncExtHidden_i; \lEncExtHidden_i]\right).
\end{align}
The extraction probability is then determined by contributions from five 
sources:
\begin{align}
    \textit{content} &\quad a^{(con)}_i=W^{(con)} z_i, \\
    \textit{salience}&\quad a^{(sal)}_i = z_i^TW^{(sal)} q, \\
    \textit{novelty}&\quad a^{(nov)}_i = -z_i^TW^{(nov)} \tanh(g_i), \label{eq:srnov} \\
    \textit{position}&\quad a^{(pos)}_i = W^{(pos)} l_i, \\
    \textit{quartile}&\quad a^{(qrt)}_i = W^{(qrt)} r_i,
\end{align}
where $l_i$ and $r_i$ are embeddings associated with the $i$-th sentence
position and the quarter of the document containing sentence $i$ respectively.
In Equation~\ref{eq:srnov}, $g_i$ is an iterative summary representation 
computed as the
sum of the previous $z_{<i}$ weighted by their extraction probabilities,
\begin{align}
g_i & = \sum_{j=1}^{i-1} p(y_j=1|y_{<j},h) \cdot z_j.
\end{align}
Note that the presence of this term induces dependence of each 
$\slabel_i$ to 
all $\slabel_{<i}$ similarly to the Cheng \& Lapata extractor.

The final extraction probability is the logistic sigmoid of the
sum of these terms plus a bias,
\begin{align}
    p(y_i=1|y_{<i}, h) &= \sigma\left(\begin{array}{l}
      a_i^{(con)} + a_i^{(sal)} + a_i^{(nov)} \\
  + a_i^{(pos)}  + a_i^{(qrt)} + b \end{array}\right).
\end{align}
The weight matrices $W_q$, $W_z$, $W^{(con)}$, $W^{(sal)}$, $W^{(nov)}$, $W^{(pos)}$,
$W^{(qrt)}$ and bias terms $b_q$, $b_z$, and $b$ are learned parameters;
The GRUs have separate learned parameters.
The hidden layer size of the GRU is 300 for each direction $z_i$, $q$, and $g_i$ have 100 dimensions. The position and quartile embeddings are 16 dimensional each.
Dropout with drop probability .25 is applied to the GRU outputs and to $z_i$.

Note that in the original paper, the SummaRunner extractor was paired 
with
an \textit{RNN} sentence encoder, but in this work we experiment with a variety
of sentence encoders.

\section{Ground Truth Extract Summary Algorithm} \label{app:oracle}

\begin{algorithm}[H]
    \DontPrintSemicolon
    \KwData{input document sentences $\sent[1], \sent[2], \ldots, \sent[n]$, 
            \\~~~~~~~~~~$\,$human reference summary $R$, 
            \\~~~~~~~~~~$\,$summary word budget $c$.}
   $y_i := 0 \quad \forall i \in 1, \ldots, n$ 
   \tcp*{Initialize extract labels to be 0.}
   $S := [\;]$ 
   \tcp*{Initialize summary as empty list.}

   \While(\tcp*[f]{While summary word count $\le$ word budget.}){$\sum_{\sent \in \summary} \textsc{WordCount}(\sent) \le c \;$}{
  ~ \\
\tcc*[l]{Add the next best sentence to the summary if it will improve the ROUGE score otherwise no improvement can be made so break.}

  ~ \\
        $\hat{i} = {\argmax}_{\substack{ i \in \{1, \ldots, n\}, \\ y_i \ne 1 }} 
        \textsc{Rouge}(\summary + [\sent[i]], R)$
        
  ~ \\
  \eIf{$\textsc{Rouge}(\summary + [ \sent[\hat{i}] ], R ) > \textsc{Rouge}(\summary, R)$}{
            $\summary := \summary + [ \sent[\hat{i}] ]$
        \tcp*{Add $s_{\hat{i}}$ to the summary sentence list.}
        $y_{\hat{i}} := 1$ 
        \tcp*{Set the $\hat{i}$-th extract label to indicate extraction.}
        }{ \textbf{break}}

    }
    
    \KwResult{extract summary labels $y_1, \ldots, y_n$}
    \caption{\textsc{OracleExtractSummaryLabels}}
\end{algorithm}

%
\hal{is this right? this could yield something longer than the budget size because what if the last step selects a really long sentence.
  i think you want the argmax over $\{j \in [n] - S : \sum \dots \leq c\}$ to make things brief maybe define $\textit{len}$ to be the current summary length? or $\textit{len}(S)$? also minor quibble but i like $S \cup \{s_i\}$ better than without the braces. CK this is what I did. I saved all summary length constraint for the eval, e.g. rouge just chops things off at 100 words. Not ideal but in practice i think it helps to have more to predict, ie less positive label sparseness, since we are no where near accurately predicting these labels anyway.}

\section{Optimizer and initialization settings.} \label{app:optset}

    We use a learning rate of .0001 and a dropout rate of .25 for all dropout
    layers. We also employ gradient clipping ($-5 < \nabla_\theta < 5$).
Weight matrix parameters are initialized using 
    Xavier initialization with the normal distribution 
    \cite{glorot2010understanding} and bias terms are set to 0.
    We use a batch size of 32 for all datasets except AMI and PubMed, which
    are often longer and consume more memory, for
    which we use sizes two and four respectively.
    For the Cheng \& Lapata model, we train for half of the maximum epochs 
    with teacher forcing, i.e. we set $p_i = 1$
    if $y_i = 1$ in the gold data and 0 otherwise 
    when computing the decoder input 
    $p_i \cdot \sentEmb[i]$; we revert to the predicted model probability 
    during the second half training.
\hal{i feel like this paragraph was produced by a summarization algorithm run on the state of Chris' brain :P --- that is to say, it's not super coherant. maybe reorder things so that they are in order of train/apply/eval or something? right now i'm not sure what matters and what doesn't matter. also so-called teacher forcing breaks the definition of log likelihood above, but maybe that's fine.}

\makeappendix

\end{document}